\documentclass[11pt,a4paper]{article}
\usepackage[hyperref]{acl2022}
\usepackage{times}
\usepackage[utf8]{vietnam}
\usepackage{latexsym}

\usepackage{graphicx}
\usepackage{booktabs}
\graphicspath{{./images/}}
\usepackage{microtype}

\aclfinalcopy 

\title{A BART-based approach with hierarchical strategy for Vietnamese abstractive multi-document summarization}

\author{Nguyen Nguyen Xuan Vu \\
  Aimesoft JSC \\
  Hanoi, Vietnam \\
  \texttt{vunx@aimesoft.com} \\\And
  Ngo Quang Huy \\
  Aimesoft JSC \\
  Hanoi, Vietnam \\
  \texttt{huynq@aimesoft.com} \\}

\date{}

\begin{document}
\maketitle
\begin{abstract}

In this technical report, we focus on solving the challenge of Vietnamese multi-document abstractive summarization, introduced in the International Workshop on Vietnamese Language and Speech Processing (VLSP) 2022. We choose to follow the popular hierarchical approach, i.e. condensing each document followed by aggregation and summarization. We propose a novel yet simple strategy to shorten documents that is driven by the golden summary, thus ensuring high correlation between stages of the hierarchical approach. Our method achieves a ROUGE2-F1 score of 0.2468 on the VLSP's public test set, and can produce fluent and concise summaries.
Additionally, we utilize external sources for extra data, which greatly enhances the quantity of data for Vietnamese multi-document summarization. The additional data is made available for the community.

\end{abstract}

\section{Introduction}
\label{sec:intro}
Living in the age of rapid technological advancement in communication and media platforms, humans often find ourselves flooded in an excessive amount of data. In order to not only capture the essential information but also envision the big picture about a topic or an event, one often has to consume numerous documents before coming up to a final conclusion. To automate this process, multi-document summarization has emerged as an attractive research topic amongst Natural Language Processing (NLP) community. In the context of this work, we will only discuss the abstractive branch of the task, which aims to generate summaries that are fluent, succinct yet able to cover principal information of a cluster of related documents. Preferably, the summaries should refrain from exactly copying text fragments from input, and instead, be paraphrased from the input contents.

The task of multi-document summarization has been extensively researched for English, yet little to no work has been done for Vietnamese. Therefore, to promote research in multi-document summary for Vietnamese, the International Workshop on Vietnamese Language and Speech Processing (VLSP) 2022 has been holding a competition dedicated to Vietnamese multi-document abstractive summarization.

For English, predominant strategies to this task include graph-based approaches \cite{liao2018abstract, li2020leveraging, pasunuru-etal-2021-efficiently}, which are efficient in capturing semantic relationships between document segments, but often have to utilize additional information, such as discourse correlation \cite{christensen2013towards}. Another approach follows a hierarchical scheme \cite{liu2019hierarchical, fabbri2019multi, jin-etal-2020-multi}, where each document in a cluster is mapped to a shorter intermediate representation, before being joined together to produce the final summary. While sounds natural, these works only generate the intermediate representation of documents based solely on the input, which may cause information shortage compared with the final summary, undesirably encouraging supervised models to make up content when generating summarization.

Simultaneously, pretrained seq-to-seq models, which are either multipurpose \cite{lewis2019bart, roberts2019exploring} or summarization-specific \cite{zhang2020pegasus}, have demonstrated the benefits of pretraining on the abstractive summarization task. Nonetheless, these models are not feasible for end-to-end multi-document summary due to the long input length. For instance, BART \cite{lewis2019bart} pretrained model can only take a token array of maximum length of 1024 as input, while an average cluster in the VLSP2022 dataset \cite{tran2022vlsp}
is approximately 2220-token long. Although recently there has been pretrained language models that are designed to tackle long input sequences \cite{guo2021longt5, xiao2021primer}, to the best of our knowledge, there has not been any Vietnamese pretrained language model that can take a sufficient number of tokens as input for multi-document tasks.



In this work, we aim to tackle the challenge of multi-document summarization for Vietnamese news using the hierarchical approach. However, differs from previous works following this scheme, we incorporate the golden summary into the process of generating the intermediate representations of documents to ensure that the aggregated intermediate representation has equal or more information compared with the final summary.

Furthermore, despite the effort, the amount of data available for Vietnamese multi-document summarization \cite{tran2020vims, tran2022vlsp} is still small with only a few hundred document clusters. Hence, to ease the training process of this task, we try to enlarge the dataset, by combining all available datasets for Vietnamese multi-document summarization along with a Vietnamese-translated version of Multinews \cite{fabbri2019multi} into one large dataset. We devote the translated Multinews dataset for the community for further research and development.

Our contributions are twofold:
\begin{itemize}
    \item We approach the challenge using the well established and effective hierarchical strategy, in which we propose a simple strategy to sample sentences from documents in a cluster that can minimize information loss for subsequent summary generation. Furthermore, our approach can be integrated well with the available seq-to-seq Vietnamese pretrained language models, such as BARTPho \cite{tran2021bartpho} or ViT5 \cite{phan2022vit5}.
    \item We translate the Multinews dataset \cite{fabbri2019multi} into Vietnamese, yielding 50286 additional document clusters and make it publicly available to community. \footnote{https://github.com/ngoquanghuy99/Abmusu2022} 
\end{itemize}
\section{Proposed Methods}
\label{sec:methods}

\subsection{Data Preprocessing}

\begin{figure}[h!]
    \centering
    \includegraphics[width=0.9\linewidth]{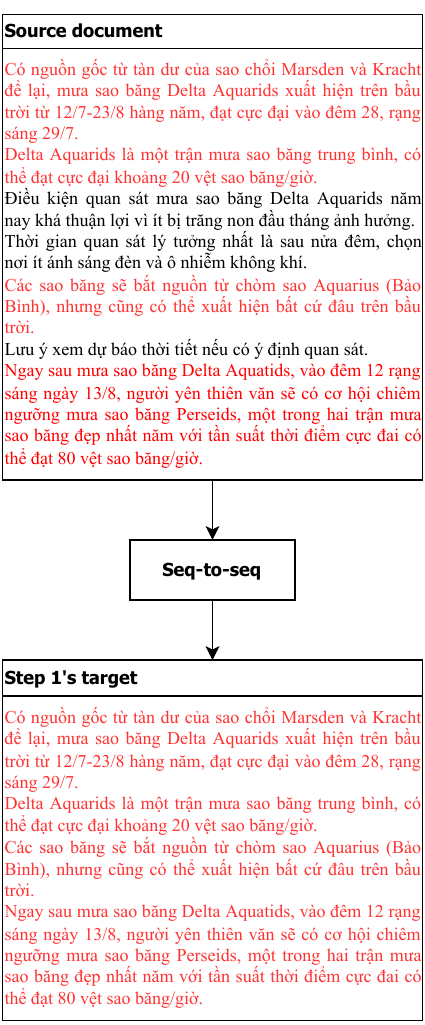}
    \caption{An example of the training objective of the first step. Important sentences are colored in red.}
    \label{fig:step1}
\end{figure}

Our approach towards the multi-document summarization problem follows the two-step regime: The first step involves shortening every document in a cluster, and in step two all shortened documents are used to infer the final summary for the whole cluster. In step one, for each cluster, we choose the top most important sentences across all documents until a desired length threshold is reached (Figure \ref{fig:step1}). As a measurement of importance, we compute ROUGE1 score \cite{lin-2004-rouge} between each sentence and the golden summary. Following the \textbf{Ind-Uniq} strategy as described in \cite{zhang2020pegasus}, we score each sentence independently and only count each \textit{n}-gram once.
Since this sentence selecting procedure is driven by the ground truth summary, a supervised model is trained to infer the selected sentences given a single document as input. 

In the second step, the chosen sentences from the first step are concatenated and fed to a seq-to-seq model to generate the final summary. To make the output summary robust to the order of the initial documents, we augment the data for the second step by permuting the order of input sentences in correspondence with the permutation of the documents' order, while keeping the summary untouched.

We argue that by selecting sentences based on the desired summary in step 1 and use these sentences for summarization in step 2, we can tackle the information mismatch between the two steps and thus prevent the model from hallucination \cite{maynez2020faithfulness} when generating summaries.

\subsection{Model Architecture}
\begin{figure}[h!]
    \centering
    \includegraphics[width=0.9\linewidth]{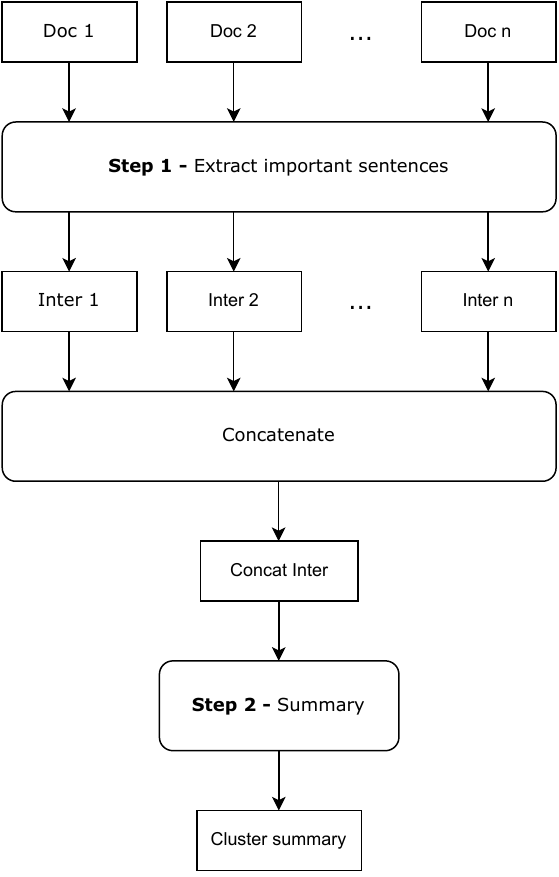}
    \caption{Flow of inference.}
    \label{fig:infer_flow}
\end{figure}

We train two separate seq-to-seq models, corresponding with the two steps described above. In step 1, the first model is trained to generate a sequence of sentences that are selected from an input document. while in step 2, the other model uses the aggregated sentences from step 1 to finally generate the summary for a cluster of documents (Figure \ref{fig:infer_flow}).
It is noted that any seq-to-seq language model can be used for both steps, including pretrained ones.

As our first attempt, we choose to experiment with BARTPho \cite{tran2021bartpho}, a BART-like model pretrained on Vietnamese corpus, due to the fact that BART was trained on a maximum input length of 1024 tokens. The long input length of BART allows us to select more sentences in step 1, thus reducing the risk of information loss when transferring between two steps.


\section{Experiments and results}
\subsection{Datasets}

The official dataset, including training, validation and test subsets, is provided by the VLSP 2022 organizer. The training set consists of 200 clusters, each cluster has the number of documents varied from 2 to 5. All documents were crawled from Vietnamese news data on a wide range of topics. To enlarge the training set, we combine the provided dataset with two other similar datasets on multi-document summarization for Vietnamese news, namely ViMs \cite{tran2020vims} and VietnameseMDS \footnote{https://github.com/lupanh/VietnameseMDS}, raising the total number of clusters up to 700 clusters. However, this number of data points is still insignificant compared with the English counterpart, Multinews \cite{fabbri2019multi}, which contains \textasciitilde56,000 clusters. Therefore, we decide to translate the Multinews dataset to Vietnamese and combine this translated set with the Vietnamese ones. In the end, we have a total \textasciitilde51,000 clusters in our final training set.

\subsection{Hyperparameters}

As mentioned, for two steps of sentence selection and summary generation, we fine-tune two distinct BARTPho \cite{tran2021bartpho} models. For both models, we set the learning rate at 5e-5, with a linear decay rate of 0.1. Due to the nature of two steps, the maximum input and output lengths of step 1's model both are set to be 512, while those of step 2's are 1024 and 256, respectively. Due to hardware limitation, the step 1's model was trained for 2,500 steps with a batch size of 2, and the other one was trained for 17,000 steps with a batch size of 4. 

During the decoding processes of both steps, we use a beam width of 3 with a length-penalty of 2.0 and a repetition penalty of 2.5. We also ensure that no 3-gram is repeated during generation.

\subsection{Experimental Results}
\subsubsection{System evaluation}

\begin{table}[]
\caption{Our method's performance on the public test set, compared with baselines provided by VLSP 2022 organizer.}
\label{tab:public_test}
\begin{tabular}{llll}
\hline
\textbf{model}       & \textbf{R1-F1} & \textbf{R2-F1} & \textbf{RL-F1} \\ \hline
\textbf{Ours}                  & 0.4550         & 0.2468         & 0.4129         \\
extractive baseline  & 0.4836         & 0.2650         & 0.4421         \\
rule baseline        & 0.4640         & 0.2582         & 0.4284         \\
anchor baseline      & 0.4381         & 0.1931         & 0.3928         \\
abstractive baseline & 0.3129         & 0.1457         & 0.2797         \\ \hline
\end{tabular}
\end{table}

Our models achieve cross-entropy losses of 1.95 and 1.97 on the validation sets of step 1 and step 2, respectively.

Aside from the training and validation set, the organization of VLSP also provides a test set, which is divided into a public set and a private set, along with four baselines.

On the public test set, our method demonstrates a relatively good performance (ROUGE2-F1=0.2468) compared with the anchor baseline (ROUGE2-F1=0.1931) and the abstractive baseline (ROUGE2-F1=0.1457) implemented by the organizer. However, we are unable to outperform the extractive (ROUGE2-F1=0.2650) and rule-based baselines (ROUGE2-F1=0.2582).

\subsubsection{Human evaluation}

Despite having a humble performance on the public test set, based on human observation, our method can produce grammatically correct and comprehensible summaries. The summaries can also mostly cover the principal content of its corresponding documents.

\section{Conclusion}
\label{sec:conclusion}

In this report, we have presented our effort on tackling the Vietnamese multi-document abstractive summarization challenge of VLSP 2022. Our proposed method shows prominent outcome and can be further tuned to produce better results. Some improvements we are thinking of include using smaller text units rather than sentences for step 1, introducing boundaries between documents when concatenating intermediate representations, and removing duplicated contents in the aggregated representation before feeding it to the step 2's model.
 
\bibliography{vlsp2022}
\bibliographystyle{acl_natbib}

\end{document}